\documentclass{article}

\usepackage{microtype}
\usepackage{graphicx}
\usepackage{subfigure}
\usepackage{booktabs} 

\usepackage{hyperref}

\usepackage[accepted]{icml2024}

\usepackage{amsmath}
\usepackage{amssymb}
\usepackage{mathtools}
\usepackage{amsthm}
\usepackage{subfigure}
\usepackage{multirow}
\usepackage{bbding}
\usepackage{float}
\usepackage{graphicx}
\usepackage[misc]{ifsym}
\usepackage{comment}
\usepackage{mmstyle}

\usepackage[capitalize,noabbrev]{cleveref}

\theoremstyle{plain}

\theoremstyle{definition}

\theoremstyle{remark}

\usepackage[textsize=tiny]{todonotes}

\icmltitlerunning{SepRep-Net: Multi-source Free Domain Adaptation via Model Separation And Reparameterization}

\begin{document}

\twocolumn[
\icmltitle{SepRep-Net: Multi-source Free Domain Adaptation via Model Separation And Reparameterization}




\begin{icmlauthorlist}
\icmlauthor{Ying Jin}{yyy}
\icmlauthor{Jiaqi Wang}{xxx}
\icmlauthor{Dahua Lin}{yyy}

\end{icmlauthorlist}

\icmlaffiliation{yyy}{Department of Information Engineering, The Chinese University of Hong Kong, Hong Kong}
\icmlaffiliation{xxx}{Shanghai AI Laboratory}

\icmlcorrespondingauthor{Jiaqi Wang}{wjqdev@gmail.com}

\icmlkeywords{Machine Learning, ICML}

\vskip 0.3in
]



\printAffiliationsAndNotice{}  

\begin{abstract}
   We consider multi-source free domain adaptation, the problem of adapting multiple existing models to a new domain without accessing the source data. 
   Among existing approaches, methods based on model ensemble are effective in both the source and target domains, but incur significantly increased computational costs.
   Towards this dilemma, in this work, we propose a novel framework called \textbf{SepRep-Net}, which tackles multi-source free domain adaptation via model \textbf{Separation} and \textbf{Reparameterization}. 
   Concretely, SepRep-Net reassembled multiple existing models to a unified network, while maintaining separate pathways (\textbf{Separation}).
   During training, separate pathways are optimized in parallel with the information exchange regularly performed via an additional feature merging unit. 
   With our specific design, these pathways can be further reparameterized into a single one to facilitate inference (\textbf{Reparameterization}). SepRep-Net is characterized by \textbf{1) effectiveness:} competitive performance on the target domain, \textbf{2) efficiency:} low computational costs, and \textbf{3) generalizability:} maintaining more source knowledge than existing solutions. 
   As a general approach, SepRep-Net can be seamlessly plugged into various methods.
   Extensive experiments validate the performance of SepRep-Net on mainstream benchmarks.
   \end{abstract}
\section{Introduction}
\label{sec:intro}

The past decade has witnessed the prosperity of deep learning. Consequently, a variety of models pre-trained on different datasets are restored to be applied. These models fall into several mainstream network architectures, such as ResNet. When facing novel datasets or tasks in real-world applications, it is important for us to make good utilization of these well-trained models, especially when data is highly limited. Therefore, in this paper, we consider adapting multiple models that are homogeneous in architecture, but pre-trained on different datasets, to a novel unlabeled dataset.

\begin{figure}[ht]
  \centering
  \subfigure[Problem Setup]{\label{fig:set}\includegraphics[width=0.28\textwidth]{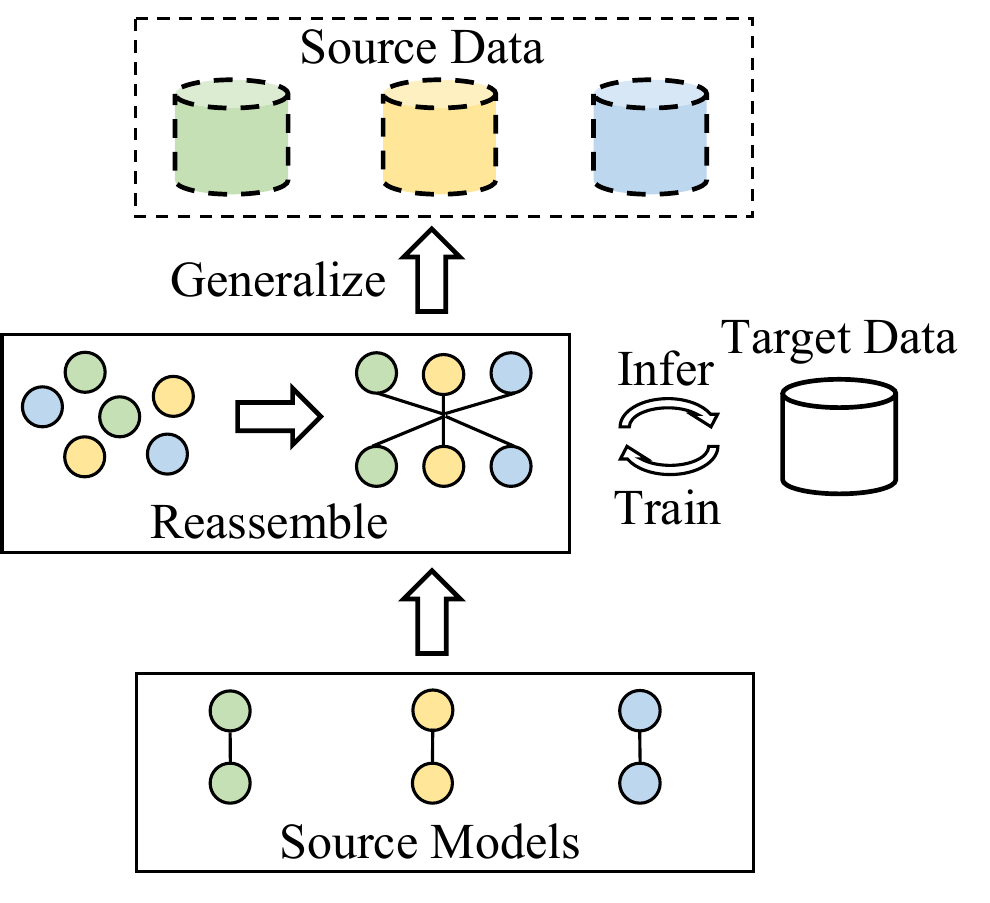}}
  \hfill
  \subfigure[Method Performance]{\label{fig:param-acc}\includegraphics[width=0.18\textwidth]{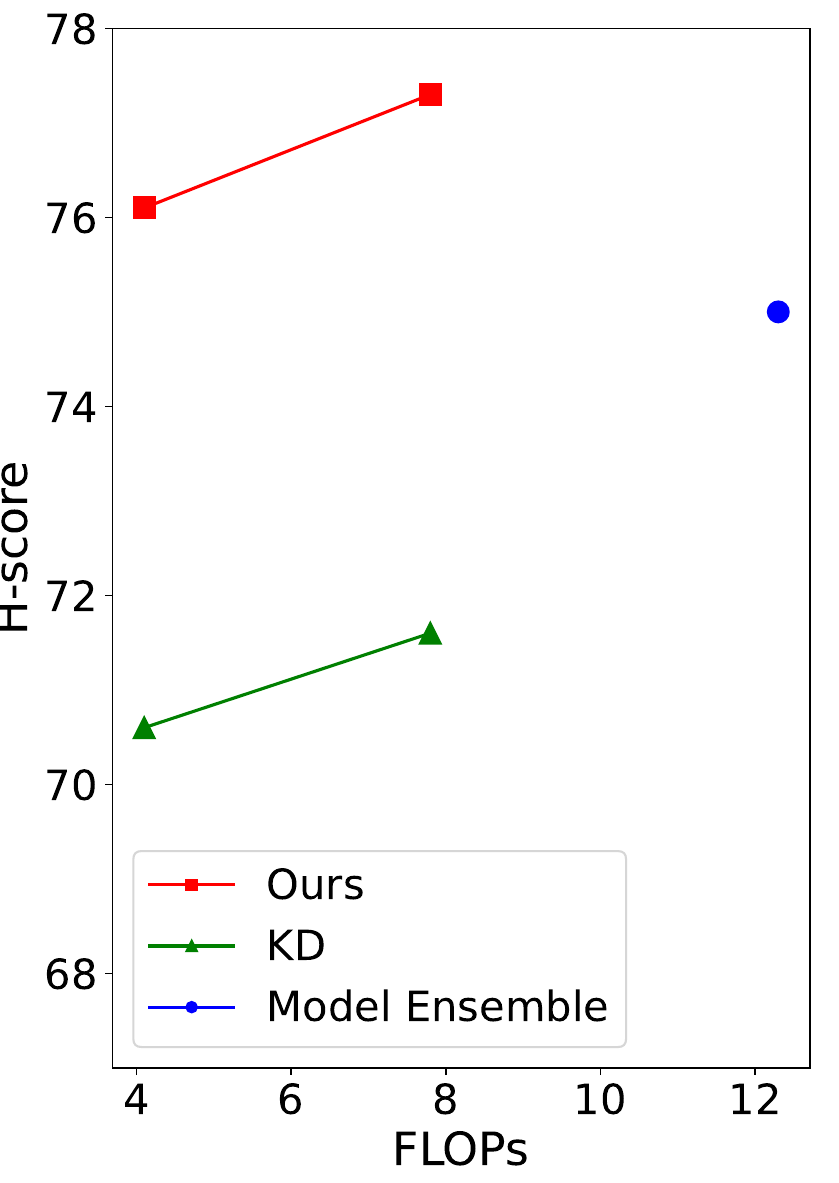}}
  \caption{\textbf{Problem Setup and Method Performance. (a)} We reassemble multiple existing source models to one target model that is adapted to a novel unlabeled target domain. The target model is expected to be 1) effective for target data, 2) efficient in inference and 3) able to generalize well on source data, \ie preserving more source knowledge. \textbf{(b)} Compared to previous methods on Office-Home, our framework enjoys better effectiveness, generalizability, and efficiency (H-score $\frac{2 \cdot Acc_{source} \cdot Acc_{target}}{Acc_{source} + Acc_{target}}$ ($\%$) evaluates effectiveness and generalizability jointly).}
  \label{fig:intro}
\end{figure}

\begin{table*}[htbp]
    \centering
    \caption{\textbf{Comparison of different domain adaptation settings.} Adapting multiple source domains to a novel target domain without source data, our method takes effectiveness, efficiency, and preserving source knowledge (\ie generalizability) into consideration.}
    \vspace{10pt}
    \resizebox*{0.8\textwidth}{!}{
        \begin{tabular}{lcccc}
            \toprule
            Method                                                    & No source data & Multiple source domains & Model efficiency & Source knowledge \\
            \midrule
            UDA~\cite{pan2010survey}                                  & \XSolidBrush   & \XSolidBrush            & -                & -                \\
            SFDA~\cite{liang2020shot}                                 & \Checkmark     & \XSolidBrush            & -                & -                \\
            MSDA~\cite{peng2019moment}                                & \XSolidBrush   & \Checkmark              & \XSolidBrush     & -                \\
            G-SFDA~\cite{yang2021generalized}                         & \Checkmark     & \XSolidBrush            & -                & \Checkmark       \\
            MSFDA~\cite{dong2021confident} & \Checkmark & \Checkmark & \XSolidBrush & \XSolidBrush \\
            \textbf{SepRep-Net (Ours)}                                & \Checkmark     & \Checkmark              & \Checkmark       & \Checkmark       \\
            \bottomrule
        \end{tabular}
    }

    \label{tab:example}
\end{table*}

Multi-source free Domain Adaptation (MSFDA) approaches can tackle the aforementioned problem, which excels at adapting multiple homogeneous models from various source datasets (domains) to a novel target unlabeled dataset. These methods are proven to be effective, enjoying outstanding performance in the target domain. However, less attention has been raised to the other side of the coin, the model accuracy in the original source datasets, which indicates the model's generability~\cite{yang2021generalized}. It is also important in some real-world applications. For example, in commercial settings, we need to adapt the models pre-trained on previous customers to the new coming customers, as well as maintain the model performance on previous customers. 

Among existing methods that pursue both effectiveness and generability~\cite{wang2020ens,decision,dong2021confident}, model ensemble~\cite{wang2020ens} serves as a simple yet strong baseline, where multiple source models are adapted to the target domain respectively.
Then these models vote for correct predictions.
These methods~\cite{decision,dong2021confident} based on model ensemble bring remarkable improvements, but they inevitably lead to another problem, high computational overheads.

Considering all the challenges mentioned above, this paper targets on a more practical setting, Multi-Source Free Domain Adaptation with Effectiveness, Efficiency, and Generalizability. As shown in Table~\ref{tab:example}, we focus on source-free domain adaptation with multiple existing source models, aiming at preserving model efficiency and source knowledge, \ie, generalizability, as well as achieving competitive target domain accuracy. To achieve this goal, an intuitive solution is to \textbf{reassemble multiple source models}, which have the same network architecture, into a single model. The reassembled model will be further adapted to the target domain. The main challenge in reassembling multiple source models lies in the trade-off between preserving more parameters and pursuing model efficiency. In this paper, we propose \textbf{SepRep-Net}  (Figure.~\ref{fig:set}), a framework for multi-source free domain adaptation via Separation and Reparameterization. For training, SepRep-Net inherits all the parameters from source models and reassembles them via Separation. While during inference, thanks to our carefully designed framework, the model can be further simplified via reparameterization. 

Concretely, SepRep-Net reassembles models by formulating multiple separate pathways in parallel (Separation). 
Specifically, in each Conv-BN unit, the input passes through separate pathways. The multiple outputs are then integrated into a single one by a feature merging unit. The unified feature serves as the input of the next Conv-BN units in different pathways. Consequently, separate pathways are optimized in parallel with information exchange regularly. More importantly, during inference, this design enables us to convert multiple pathways into a single one via reparameterization to enhance model efficiency. In other words, we eventually convert multiple source models into a single target model.
Moreover, when ensembling final classifier outputs, previous works~\cite{decision} adopt learnable combination weights, resulting in bias to target data. 
In this paper, we revise the reweighting strategy to an uncertainty-based one to further boost generalizability. 
Our method is characterized by 1) Effectiveness: reassembled model is ready to be adapted to the target data, 2) Efficiency: largely reduces computational costs during inference via reparameterization, 3) Generalizability: parameters inherited from multiple source models naturally preserve source knowledge.

Our proposed SepRep-Net can be readily integrated into various existing methods.
Extensive experiments are performed on several benchmarks to evaluate SepRep-Net. 
Experimental results (as illustrated in Figure~\ref{fig:param-acc} as an example) show that SepRep-Net 
achieves a better trade-off among effectiveness, efficiency, and generalizability.

\section{Related Work}
\label{sec:related}
\paragraph{Domain Adaptation} 
Aiming at mitigating the distribution gap between the source and target domains, mainstream domain adaptation methods mostly fall into two paradigms, \ie, moment matching~\cite{MKMMD2012,Long15DAN,Tzeng14DDC,JAN2017,CAN2019,SWD2019,xiao2023spa} and adversarial training~\cite{DANN2016,ADDA2017,Pei18MADA,Long18CDAN,Hoffman18CyCADA,symDA2019}. The former alleviates domain shift by minimizing feature discrepancy while the latter borrows the spirit of Generative Adversarial Networks (GANs)~\cite{goodfellow2014generative,CGAN2014} to learn domain invariant features. Recent works enlighten some novel perspectives for domain adaptation, such as clustering~\cite{TPN2019,tang2020unsupervised}, self-training~\cite{DIRT,CBST,Zou2019,jin2020minimum,saito2020universal}, network architecture design~\cite{carlucci2017autodial,li2018adaptive,TransNorm} and feature norm~\cite{AFN2019}. In order to leverage the source knowledge effectively, accessing source data during training is necessary for these methods, which may be unavailable under some circumstances.

\paragraph{Source-free Domain Adaptation} Previous researches focuses on transferring or training models with limited labeled data~\cite{xuhong2018explicit,li2018delta,cao2021few,jin2020transferring,zhou2023improving,jin2022semi}. Adapting source models to a novel unlabeled target domain without accessible source data, namely Source-free Domain Adaptation~\cite{liang2020shot} (SFDA), is a highly practical problem. Recent works have explored diverse scenarios in this field, \eg closed-set~\cite{liang2020shot,li2020model,liang2021source,yang2020unsupervised}, open-set~\cite{kundu2020towards} and universal SFDA~\cite{kundu2020universal}.
Our work is related to multi-source SFDA~\cite{decision} (MSFDA) and Generalized SFDA~\cite{yang2021generalized} (G-SFDA), which pay attention to adapting multiple models to the novel target domain and the model performance on source domains, \ie maintaining source knowledge, respectively.

\paragraph{Multi-source Free Domain Adaptation} Previous works~\cite{xu2018deep,peng2019moment,wang2020learning,yang2020curriculum,Nguyen2021} widely investigate multi-source domain adaptation~\cite{sun2015survey} (MSDA) which adapts multiple source domains with different distributions. When MSDA meets the source-free setting, the problem becomes multi-source free domain adaptation (MSFDA). Model ensemble works as a simple yet effective baseline. Recent advances~\cite{decision,dong2021confident} push it to a stronger level but still suffer from heavy computational costs.
To reduce the computation overheads, knowledge distillation (KD) is adopted as an effective solution. It preserves the target performance but leads to severe source knowledge forgetting, harming the generalizability.
\section{Methodology}
\label{sec:method}
This work aims at proposing a framework for adapting multiple existing models from different source domains to a novel target domain without accessing source data.
In this setting, we are given $K$ source models pre-trained on different source data and $n_{t}$ unlabeled samples $\mathcal{X}_{t} = \{x_{t}^{i}\}_{i=1}^{n_{t}}$ from the target domain.
Both the source domains and the target domain have $C$ categories. Thus, it is a $C$-way classification task. The $k^{th}$ source model consists of one feature extractor $g_{s}^{k}$ and one classifier $h_{s}^{k}$. And all source models
are homogeneous in network architecture. Our goal is to reassemble these multiple source models to a new model on the target domain using unlabeled target data, pursuing effectiveness, efficiency, and generalizability. 

To smooth the presentation of our framework, we start from the preliminaries. 
Then, we propose SepRep-Net, an efficient framework for multi-source free domain adaption.

\subsection{Preliminaries}

\paragraph{Model Ensemble} We target on source-free domain adaptation (SFDA) from multiple models, which is also coined Multi-Source Free Domain Adaptation (MSFDA). Some methods are designed specifically for this problem and proved to be effective. Except for them, another simple yet robust solution for this problem is model ensemble, where we utilize the existing SFDA methods (\eg SHOT~\cite{liang2020shot}) to adapt each source model to the target data.
As a result, we obtain $K$ adapted target models $\{g_{t}^{k}, h_{t}^{k}\}_{k=1}^{K}$ and can ensemble predictions from these adapted models as the final result,
\begin{equation}
  \hat{y}_{t}=\sum_{k=1}^{K}{\alpha_{k}\cdot h_{t}^{k}(g_{t}^{k}(x_{t}))} ,
\end{equation}
where $\alpha_{k}$ denotes the importance weight of the $k^{th}$ model. We may simply take $\alpha_{k}=\frac{1}{K}$. Though effective, model ensemble passes each data sample through all $K$ models in inference, bringing around $K$ times computational and parameter costs.

\begin{figure*}
  \centering
  \includegraphics[width=0.75\textwidth]{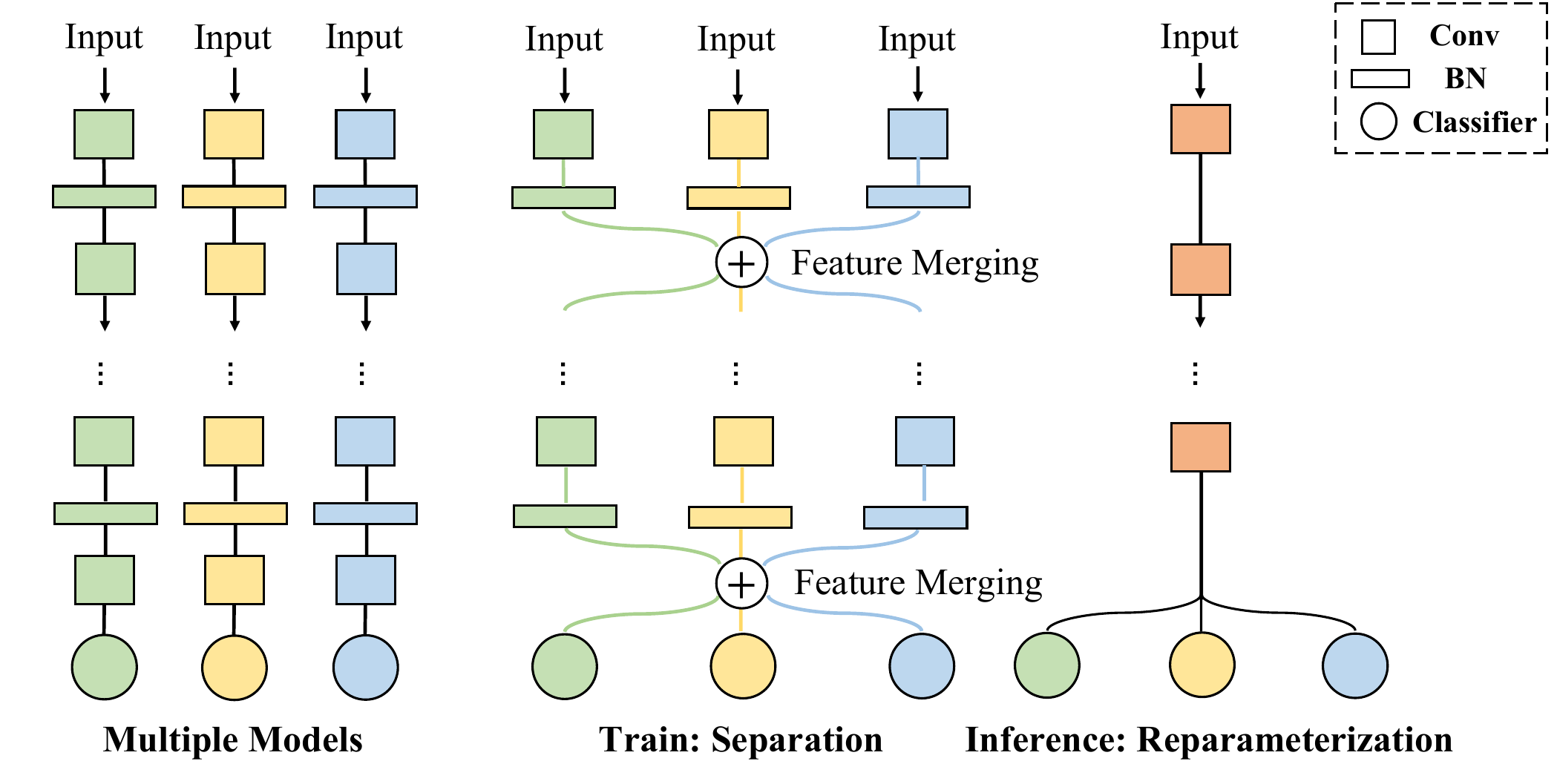}
  \caption{\textbf{Method Overview.} Take three source models as an example. \textbf{Train:} In each Conv-BN unit, separate pathways are applied in parallel, forming a structure with multiple pathways. The outputs are then integrated into a unified output via the feature merging unit. \textbf{Inference:} Multiple pathways are re-parameterized to a unique one during inference. An uncertainty-based weighting strategy ensembles multiple classifier heads to obtain the final prediction. \textit{(Best viewed in color)}}
  \label{fig:arch}
  \vspace{-10pt}
\end{figure*}

\subsection{SepRep-Net}

In this paper, we propose a framework for integrating multiple source models, namely SepRep-Net, to boost efficiency, generalizability, as well as effectiveness. As shown in Figure.~\ref{fig:arch}, SepRep-Net reassembles multiple source models to a single model by formulating separate pathways during training (Separation). In each Conv-BN unit, the input data passes through multiple pathways, and then the corresponding outputs are merged into a single one. During inference, such a design enables us to convert the multiple pathways to a single one via reparameterization. 

\paragraph{Train: Separation} 

During the training period, SepRep-Net inherits the parameters from source models and reassembles them into one network. As shown in Figure.~\ref{fig:arch}, the weights from source models (square: Conv, rectangle: BN) form multiple separate pathways for model forward. Afterward, the outputs from different pathways are converted into a unified one via the feature merging unit, which integrates them via weighted sum. Concretely, take one Conv-BN unit as an example, denote the input as $f^{input}$ and the corresponding output as $f^{output}$, we have
\begin{equation}\label{Eq:reassem}
  f^{output} = \sum_k BN_k(Conv_k(f^{input})) \cdot w_k,
\end{equation}
where $Conv_{k}$ and $BN_k$ indicates the convolution and normalization layer from the $k^{th}$ model respectively, and $w_k$ is the weight hyper-parameter of the $k^{th}$ model. In our experiments, we take $w_k = \frac{1}{K}$.
Through this framework, we can obtain multiple classifier outputs.

Next, we need to ensemble these classifier outputs. Here, we consider that different pathways cannot be viewed as equal during training. Therefore, a re-weighting strategy shall be introduced. Concretely, we state that the models with lower uncertainty on the target data are more likely to have stronger performance. To highlight these models, we first quantify the model uncertainty on the target data by Entropy.  
\begin{equation}
  M(g_{s}^{k}, h_{s}^{k}) = -\mathbb{E}_{x_t \epsilon \mathcal{X}_t} \sum_{c=1}^{C} h_s^k(g_s^k(x_t))^c log(h_s^k(g_s^k(x_t))^c),
  \label{Eq:entropy}
\end{equation}
where $h_s^k(\cdot)^c$ indicates the prediction on the $c^{th}$ category of classifier $h_s^k$, and $M(g_{s}^{k}, h_{s}^{k})$ is a score that indicates the prediction uncertainty of the $k^{th}$ source model on target data $x_t$. A lower score demonstrates lower prediction uncertainty, which shows that the model is potentially more capable of adapting to the target data.  

Our proposed SepRep-Net leverages the uncertainty metric in Eq.~\ref{Eq:entropy} with Softmax function to re-weight the losses on different classifiers, 
\begin{equation}\label{Eq:loss}
  L_{total} = \sum_{k=1}^{K} \frac{e^{-M(g_{s}^{k}, h_{s}^{k})}}{\sum_{i=1}^{K}e^{-M(g_{s}^{i}, h_{s}^{i})}} L_k(f_k; \mathcal{X}_t),
\end{equation}
where $L_k(f_k; \mathcal{X}_t)$ denotes the loss function of the $k^{th}$ model. The loss functions~\cite{liang2020shot} in previous works for SFDA can be readily served as $L_k$ in SepRep-Net.  

\paragraph{Inference: Reparameterization}
For inference, as shown in Figure~\ref{fig:arch}, our framework design enables us to merge these separate pathways into a unified one with model reparameterization, which enhances the model efficiency. 

Concretely, take one Conv-BN unit as an instance, suppose one input feature feeds to this unit is $x^{(1)} \in \mathbb{R}^{H_1\times W_1\times C_1}$, where $H_1 \times W_1$ shows the spatial resolution and $C_1$ is the channel size. For the convolution layer, take the one in the $k^{th}$ pathway as an instance, suppose there are $C_2$ convolution filters in total, the $j^{th}$ filter can be denoted as $F^{(k,j)}\in\mathbb{R}^{U\times V\times C_1}$, the corresponding output of this convolution layer $x^{(2)} \in\mathbb{R}^{H_2\times W_2\times C_2}$ follows
\begin{equation}\label{eq-def-conv}
  x^{(2,k)}_{:,:,j}=\sum_{i=1}^{C_1} x^{(1)}_{:,:,i}\ast F^{(k, j)}_{:,:,i},
\end{equation}
where $x^{(1)}_{:,:,i}$ is the $i^{th}$ channel of $x^{(1)}$, $F^{(k, j)}_{:,:,i}$ is the $i^{th}$ channel of $F^{(k, j)}$, and $\ast$ is the 2D convolution operator. Afterward, it passes through the batch normalization layer in this pathway, and the output $o \in\mathbb{R}^{H_2\times W_2\times C_2}$ becomes
\begin{equation}\label{eq-conv-with-bn}
  o_{:,:,j}^{k}=(x^{(2,k)}_{:,:,j} - \widetilde{\mu}^k_j)\frac{\gamma^k_j}{\widetilde{\sigma}^k_j} + \beta^k_j \,,
\end{equation}
where $\widetilde{\mu}^k_j$ and $\widetilde{\sigma}^k_j$ are the $j^{th}$ element of the mean and standard deviation statistics in batch normalization, $\gamma^k_j$ and $\beta^k_j$ are the $j^{th}$ element of the learned scaling factor and bias respectively.
Therefore, the weighted sum of the outputs of multiple pathways is
\begin{equation}\label{eq-output}
  x^{(3)}_{:,:,j}=\sum_k w_k \cdot ((\sum_{i=1}^{C_1} x^{(1)}_{:,:,i}\ast F^{(k, j)}_{:,:,i} - \widetilde{\mu}^k_j)\frac{\gamma^k_j}{\widetilde{\sigma}^k_j} + \beta^k_j).
\end{equation}

We can reparameterize the output as follows,

\begin{equation}\label{eq-fused-kernel}
  F^{\prime(j)} = \frac{w_1 \gamma^1_j}{\widetilde{\sigma}^1_j}F^{(1, j)} \oplus \frac{w_2 \gamma^2_j}{\widetilde{\sigma}^2_j}F^{(2, j)} \oplus ... \oplus \frac{w_K \gamma^K_j}{\widetilde{\sigma}^K_j}F^{(K, j)} \,,
\end{equation}
\begin{equation}\label{eq-fused-bias}
  b_j = - w_1 \frac{\widetilde{\mu}^1_j \gamma^1_j}{\widetilde{\sigma}^1_j} - ... - - w_K \frac{\widetilde{\mu}^K_j \gamma^K_j}{\widetilde{\sigma}^K_j} + w_1 \beta^1_j + ... + w_K \beta^K_j \,,
\end{equation}
\begin{equation}\label{eq-fused}
  x^{(3)}_{:,:,j} =\sum_{i=1}^{C_1} x^{(1)}_{:,:,i} \ast F^{\prime(j)}_{:,:,i} + b_j ,
\end{equation}
where $\oplus$ indicates element-wise addition on the corresponding positions. Now, the multiple pathways in one Conv-BN unit, with $K$ convolution layers and $K$ batch normalization layers, are converted to one convolution layer, as shown in Eq.~\ref{eq-fused}.

As shown in Figure.~\ref{fig:arch}, multiple models are finally reassembled into a unified architecture with multiple classifier heads.
Since the parameter and computation overheads of classifier heads are negligible compared to the feature extractor, the computation and parameter costs of SepRep-Net are very close to a single model.

\paragraph{Importance Reweighting} Finally, akin to the spirit of model ensemble, we take the weighted sum of multiple classifiers as the final prediction,
\begin{equation}\label{eq-output}
  \hat{y}_{t}=\sum_{k=1}^{K}{\alpha_{k}\cdot h_{t}^{k}(g_t(x_{t}))},
\end{equation}
where $\alpha_k$ is the ensemble weight which plays an important role in model ensemble. A recent work~\cite{decision} proposed to learn a combination weight during the training procedure. However, such a strategy inevitably causes bias to target data, which hurts the model generalizability. Here, we revise the reweighting strategy to a parameter-free one. We take the entropy of the trained models to formulate our reweighting strategy, through which we can highlight the model that is more certain on the test data,
\begin{equation}
  \alpha_k = \frac{e^{-M(g_t, h_{t}^{k})}}{\sum_{i=1}^{K}e^{-M(g_{t}, h_{t}^{i})}}.
\end{equation}
Notably, $M(g_t, h_{t}^{k})$ is different from the entropy criterion $M(g_{s}^{k}, h_{s}^{k})$ in Eq.~\ref{Eq:entropy} since 1) it is calculated by adapted model instead of the source models to obtain a more accurate approximation of uncertainty, 2) it is computed on-the-fly in inference, harvesting importance weights that are adaptive to the input data. 

\section{Experiments}
\label{sec:experiment}

\subsection{Experimental Setup}
\paragraph{Datasets} We perform an extensive evaluation of SepRep-Net on five mainstream benchmark datasets, \ie, \textbf{Office-31}~\cite{Saenko10Office}, \textbf{Office-Home}~\cite{Venkateswara17Officehome}, \textbf{Digit5}~\cite{peng2019moment}, \textbf{Office-Caltech}~\cite{GFK}, and \textbf{DomainNet}~\cite{peng2019moment}.
Due to the page limit, detailed descriptions of the datasets are included in Appendix.

\paragraph{Implementation Details} Following previous works, we take ResNet pre-trained on ImageNet as the feature extractor to recognize objects in the real world. For digit recognition, we adopt the network structure in \cite{liang2020shot} and train it from scratch. We use the same bottleneck layer and task-specific classifier as \cite{liang2020shot}. We take $w_k = \frac{1}{K}$ in Eq.~\ref{Eq:reassem} to reassemble the multiple networks.

\subsection{Experimental Results}

\begin{table*}[ht]
    \centering
    \caption{Source Accuracy (S) (\%), Target Accuracy (T) (\%), H-Score (H), and Model Efficiency on Office-31 dataset. ResNet-50 is adopted in experiments. A, D, W indicate different domains (A: Amazon, D: DSLR, W: Webcam). 
    SHOT-ens indicates the performance of model ensemble with all models that are adapted via SHOT. KD indicates knowledge distillation. 
    }
    \vspace{10pt}
    \small
    \resizebox*{0.8\textwidth}{!}{
        \begin{tabular}{l|c|ccc|ccc|ccc|ccc}
            \toprule
            \multirow{2}{*}{METHOD}  & \multirow{2}{*}{FLOPS} & \multicolumn{3}{c|}{D, W $\rightarrow$ A} & \multicolumn{3}{c|}{A, W $\rightarrow$ D} & \multicolumn{3}{c|}{A, D $\rightarrow$ W} & \multicolumn{3}{c}{Average} \\
            &           & S      & T      & H      & S      & T      & H      & S      & T      & H      & S      & T      & H   \\
            \midrule
            DECISION + KD~\cite{hinton2015distilling} & 4.1 & 73.4 & 75.4& 74.4 & 72.5 & 99.6 & 83.9  & 76.7  & 98.1 & 86.1 & 74.2 & 91.0 & 81.4 \\
            DECISION \textbf{+ SepRep-Net}  & 4.1 & 87.5  & 77.1  & 81.3 & 90.3 & 99.7  & 94.8 & 88.6 & 99.1 & 93.3 & \textbf{88.8} & \textbf{92.0} & \textbf{90.4} \\
            \midrule
            DECISION~\cite{decision} & 8.2 & 92.8 & 75.4 & 83.2 & 81.9 & 99.6 & 89.8 & 83.1 & 98.4 & 90.1 & 85.9 & 91.1 & 88.4 \\
            \midrule
            \midrule
            CAiDA + KD~\cite{hinton2015distilling}  & 4.1 & 84.2 & 75.7 & 79.7 & 85.1 & 99.4 & 91.7 & 84.7 & 98.7 & 91.2 & 84.5 & 91.3 & 87.8 \\
            CAiDA + \textbf{SepRep-Net}  & 4.1 & 92.0  & 76.4  & 83.5 & 89.3 & 99.7  & 94.2 & 89.2 & 99.3  & 94.0 & \textbf{90.2} & \textbf{91.8} & \textbf{91.1} \\
            \midrule
            CAiDA~\cite{dong2021confident}     & 8.2  & 88.9 & 75.8 & 81.8 & 89.1 & 99.8 & 94.1 & 88.6 & 98.9 & 93.5 & 88.9 & 91.6 & 90.2 \\
            \midrule
            \midrule
            SHOT-ens + KD~\cite{hinton2015distilling}  & 4.1 & 85.3 & 74.9 & 79.8 & 86.4 & 97.8 & 91.7 & 85.8 & 94.8 & 90.1 & 85.8 & 89.2 & 87.5\\
            SHOT + \textbf{SepRep-Net}  & 4.1 & 92.8 & 75.7 & 83.4 & 93.9 & 98.8 & 96.3 & 91.4 & 96.6 & 93.9 & \textbf{92.7} & \textbf{90.4} & \textbf{91.5} \\
            \midrule
            SHOT-ens~\cite{liang2020shot}     & 8.2  & 90.0 & 75.0 & 81.8 & 90.2 & 97.8 & 93.8 & 90.5 & 94.9 & 92.6 & 90.2 & 89.3 & 89.7 \\
            \bottomrule
        \end{tabular}
    }
    \label{tab: Acc-Office}
\end{table*}

\begin{table*}[ht]
    \centering
    \caption{Source Accuracy (S) (\%), Target Accuracy (T) (\%), H-Score (H), and Model Efficiency on Office-Home dataset. ResNet-50 is adopted in experiments. Ar, Cl, Pr, Rw indicate different domains (Ar: Art, Cl: Clipart, Pr: Product, Rw: Real World). The abbreviations in METHOD are the same as Table~\ref{tab: Acc-Office}.}
    \vspace{10pt}
    \label{tab: Acc-Office-home}
    \small
    \resizebox*{0.95\textwidth}{!}{
        \begin{tabular}{l|c|ccc|ccc|ccc|ccc|ccc}
            \toprule
            \multirow{2}{*}{METHOD}  & \multirow{2}{*}{FLOPS}   & \multicolumn{3}{c|}{Cl,Pr,Rw $\rightarrow$ Ar} & \multicolumn{3}{c|}{Ar,Pr,Rw $\rightarrow$ Cl} & \multicolumn{3}{c|}{Ar,Cl,Rw $\rightarrow$ Pr} & \multicolumn{3}{c|}{Ar,Cl,Pr $\rightarrow$ Rw} & \multicolumn{3}{c}{Average}  \\                  
            &                        & S      & T      & H      & S      & T      & H      & S      & T      & H      & S      & T      & H   & S      & T      & H \\             
            \midrule
            \midrule
            DECISION + KD~\cite{hinton2015distilling}  & 4.1  & 57.2  & 74.4 & 64.7  & 51.4   & 59.1  & 55.0 & 52.7  & 84.4 & 64.9 & 59.6 & 83.7 & 69.6   & 55.2  & 75.4 & 63.5  \\
            DECISION + \textbf{SepRep-Net}  & 4.1 & 73.2 & 76.0 & 74.6 & 69.5 & 61.6 & 65.2 & 68.8 & 85.4 & 76.3 & 71.8 & 85.2 & 78.0 & \textbf{70.8} & \textbf{77.1} & \textbf{73.9}\\
            \midrule
            DECISION~\cite{decision} & 12.3 & 68.7 & 74.5 & 71.5 & 57.3 & 59.4 & 58.3 & 67.7 & 84.4 & 75.1 & 66.8 & 83.6 & 74.3 & 65.1 & 75.4 & 69.9 \\ 
            \midrule
            \midrule
            CAiDA + KD~\cite{hinton2015distilling} & 4.1 & 68.8 & 75.0 & 71.8 & 63.4 & 60.4 & 61.9 & 66.6 & 84.5 & 74.5 & 66.0 & 84.0 & 73.9 & 66.2 & 76.0 & 70.8 \\
            CAiDA + \textbf{SepRep-Net}  & 4.1   & 75.2  & 76.3 & 75.7 & 73.6 & 61.7 & 67.0 & 75.0 & 86.3  & 80.3 & 74.8 & 85.8 & 80.0 & \textbf{74.7} & \textbf{77.5}  & \textbf{76.1} \\
            \midrule
            CAiDA~\cite{dong2021confident}   & 12.3 &  75.0  & 75.2 & 75.1  &  73.8 & 60.5 & 66.5  &  74.5 & 84.7 & 79.3  &  73.9 & 84.2 &  78.7 & 74.3 & 76.2 &  75.2 \\
            \midrule
            \midrule
            SHOT-ens + KD~\cite{hinton2015distilling} & 4.1 & 69.0 & 72.1 & 70.5 & 65.6 & 59.0 & 62.1 & 67.9 & 82.7 & 74.6 & 66.5 & 82.9 & 73.8 & 67.3 & 74.2 & 70.6 \\
            SHOT + \textbf{SepRep-Net}  & 4.1 & 82.6 & 72.3 & 77.1 & 75.6 & 59.2 & 66.4 & 75.1 & 83.3 & 79.0 & 77.8 & 83.3 & 80.5 & \textbf{77.8} & \textbf{74.5} & \textbf{76.1}  \\
            \midrule
            SHOT-ens~\cite{liang2020shot}   & 12.3 &  76.5  & 72.2 &  74.3 &  75.5 & 59.3 &  66.4 &  75.4 & 82.8 &  78.9 &  75.3 & 82.9 &  78.9 &  75.7 & 74.3 &  75.0                \\
            \bottomrule
        \end{tabular}
    }
\end{table*}

\begin{table*}[ht]
  \centering
  \caption{Source Accuracy (S) (\%), Target Accuracy (T) (\%), H-Score (H) and Model Efficiency on Office-Caltech dataset. ResNet-50 is adopted in experiments. A, C, D, W indicate different domains (A: Amazon, C: Caltech, D: DSLR, W: Webcam). SHOT-ens indicates the performance of model ensemble with all models that are adapted via SHOT. KD indicates knowledge distillation.}
  \vspace{10pt}
  \resizebox*{0.85\textwidth}{!}{
  \begin{tabular}{l|c|ccc|ccc|ccc|ccc|ccc}
    \toprule
    \multirow{2}{*}{METHOD}  & \multirow{2}{*}{FLOPS}   & \multicolumn{3}{c|}{C,D,W $\rightarrow$ A} & \multicolumn{3}{c|}{A,D,W $\rightarrow$ C} & \multicolumn{3}{c|}{A,C,W $\rightarrow$ D} & \multicolumn{3}{c|}{A,C,D $\rightarrow$ W} & \multicolumn{3}{c}{Avg}  \\                  
    &      & S      & T      & H      & S      & T      & H      & S      & T      & H      & S      & T      & H   & S      & T      & H \\           
    \midrule
    ResNet~\cite{he2016Resnet}   & 12.3 & -  & 88.7  & - & -  & 85.4 & - & - & 98.2 & - & - & 99.1 & - & - & 92.9 & - \\
    DAN~\cite{Long15DAN}   & 12.3  & - & 91.6 & - & - & 89.2 & - & - & 99.1 & - & - & 99.5 & - & -  & 94.8  & - \\
    DCTN~\cite{xu2018deep}   & 12.3 & - & 92.7 & - & - & 90.2 & - & - & 99.0 & - & - & 99.4  & - & - & 95.3 & - \\
    MCD~\cite{MCD2018}    & 12.3 & - & 92.1 & - & - & 91.5 & - & - & 99.1 & - & - & 99.5 & - & - & 95.6 & - \\
    M3SDA~\cite{peng2019moment}   & 12.3 & - & 94.5 & - & - & 92.2  & - & - & 99.2 & - & - & 99.5 & - & - & 96.4  & - \\
    \midrule
    DECISION + KD~\cite{hinton2015distilling} & 4.1 & 88.3 & 96.0 & 92.0 & 90.7 & 95.7 & 93.1 & 87.9 & 99.4 & 93.3 & 85.0 & 99.7 & 91.8 & 88.0  & 97.7 & 92.5 \\
    DECISION + \textbf{SepRep-Net} & 4.1 & 94.8 & 96.1 & 95.4 & 95.1 & 96.1 & 95.6 & 90.2 & 100.0 & 95.6 & 92.5 & 99.8 & 96.0 & \textbf{93.2} & \textbf{98.0} & \textbf{95.5}\\
    \midrule
    DECISION~\cite{decision} & 12.3 & 93.1 & 95.9 & 94.5 & 93.1 & 95.9 & 94.5& 88.1 & 100.0 &93.7 & 91.4 & 99.6 & 95.3 & 91.4 & 98.0 & 94.6\\
    \midrule
    \midrule
    CAiDA + KD~\cite{hinton2015distilling} & 4.1 & 91.5 & 96.6 & 94.0 & 92.3 & 96.8 & 94.5 & 94.6 & 100.0 & 97.2 & 95.9 & 99.8 & 97.8 & 93.6 & 98.3 & 95.9 \\
    CAiDA + \textbf{SepRep-Net}  & 4.1 & 97.2 & 97.1 & 97.1  & 96.8 & 97.1  & 96.9 & 98.4 & 100.0  & 99.2 & 97.9 & 99.7 & 98.8 & \textbf{97.6} & \textbf{98.5} & \textbf{98.0} \\
    \midrule
    CAiDA~\cite{liang2020shot}  & 12.3  & 96.0 & 96.8 & 96.4 & 96.2 & 97.1 & 96.6 & 98.0 & 100.0 & 99.0 & 98.1 & 99.8 & 98.9 & 97.1 & 98.4  & 97.7 \\
    \midrule
    \midrule
    SHOT-ens + KD~\cite{hinton2015distilling} & 4.1 & 93.4 & 95.6 & 94.5 & 95.0 & 95.7 & 95.3 & 94.2 & 96.6 & 95.4& 90.3 & 96.5 & 93.3 & 93.2 & 96.1 & 94.6 \\
    SHOT + \textbf{SepRep-Net}  & 4.1 & 99.3 & 95.9 & 97.6  & 98.1 & 95.6  & 96.8 & 97.7 & 97.5  & 97.6 & 98.1 & 99.7 & 98.9 & \textbf{98.3} & \textbf{97.2}  & \textbf{97.7} \\
    \midrule
    SHOT-ens~\cite{liang2020shot}  & 12.3  & 98.7 & 95.7 & 97.2 & 98.0 & 95.8 & 96.9 & 98.3 & 96.8 & 97.5 & 98.1  & 99.6 & 98.8 & 98.3 & 97.0  & 97.6 \\

    \bottomrule
  \end{tabular}
  }
  \label{tab: Acc-Office-caltech}
\end{table*}

\begin{table*}[ht]
    \centering
    \caption{Source Accuracy (S) (\%), Target Accuracy (T) (\%), H-Score (H), and Model Efficiency on Digit5. The backbone network in \cite{peng2019moment} is adopted to recognize digits. MM, MT, UP, SV, and SY indicate different domains (MM: MNIST-M, MT: MNIST, UP: USPS, SV: SVHN, SY: Synthetic Digits). $\rightarrow$ points to the target domain while the remaining domains serve as source domains. The abbreviations in METHOD are the same as Table~\ref{tab: Acc-Office}, and traditional domain adaptation methods that require source data (from DAN to M3SDA) are also compared.}
    \label{tab: Acc-digit}
    \footnotesize
    \vspace{10pt}
    \resizebox*{0.95\textwidth}{!}{
        \begin{tabular}{l|c|ccc|ccc|ccc|ccc|ccc|ccc}
            \toprule
            \multirow{2}{*}{METHOD} & \multirow{2}{*}{FLOPS} & \multicolumn{3}{c|}{ $\rightarrow$ MM} & \multicolumn{3}{c|}{$\rightarrow$ MT} & \multicolumn{3}{c|}{ $\rightarrow$ UP} & \multicolumn{3}{c|}{$\rightarrow$ SV} & \multicolumn{3}{c|}{$\rightarrow$ SY} & \multicolumn{3}{c}{Avg} \\
            &                        & S      & T      & H      & S      & T      & H      & S      & T      & H      & S      & T      & H   & S      & T      & H      & S      & T      & H\\
            \midrule
            DAN~\cite{Long15DAN}   & 0.116 &  -&  63.7 &  -&  -& 96.3 &  -&  - & 94.2 &  -&  -& 62.5 &  -&  - & 85.4 &  -&  - & 80.4  &  -\\
            DANN~\cite{DANN2016}  & 0.116  &  -&  71.3 &  -&  -& 97.6 &  -&  -& 92.3 &  -&  - & 63.5 &  -&  - & 85.3 &  -&  -& 82.0 &  -\\
            MCD~\cite{MCD2018}   & 0.116 &  -&  72.5 &  -&  - & 96.2 &  -&  - & 95.3  &  -&  - & 78.9 &  -&  - & 87.5 &  -&  -& 86.1 &  -\\
            CORAL~\cite{coral2016}   & 0.116 &  -&  62.5 &  -&  - & 97.2 &  -&  -& 93.4 &  -&  - & 64.4  &  -&  - & 82.7  &  -&  - & 80.1 &  -\\
            ADDA~\cite{ADDA2017}   & 0.116 &  -&  71.6 &  -&  -& 97.9 &  -&  -& 92.8  &  -&  - & 75.5 &  -&  -& 86.5 &  -&  - & 84.8  &  -       \\
            M3SDA~\cite{peng2019moment}    & 0.116 &  -&  72.8 &  -&  - & 98.4 &  -&  -& 96.1 &  -&  -& 81.3 &  -&  - & 89.6  &  -&  -& 87.6  &  -\\
            \midrule
            \midrule
            DECISION + KD~\cite{hinton2015distilling}  & 0.029 & 55.3 & 92.8 & 69.3    & 52.3   & 99.2  & 68.5  & 46.6 & 97.8 & 63.1  & 52.4  & 82.6 & 64.1  & 70.9  & 97.2 & 82.0 & 55.5  & 93.9 & 69.4  \\
            DECISION + \textbf{SepRep-Net}  & 0.032  & 61.8 & 93.4 & 74.4 & 57.0 & 99.4 & 72.5 & 65.8 & 98.3 & 78.8 & 62.1 & 84.2 & 71.5 & 73.4 & 97.7 & 83.8 & \textbf{64.0} & \textbf{94.6} & \textbf{76.4}\\
            \midrule
            DECISION~\cite{decision} & 0.116 & 58.4 & 93.0 & 71.7 & 55.1 & 99.2 & 70.8 & 55.8 & 97.8 & 71.1 & 54.0 & 82.6 & 65.3 & 71.0 & 97.5 & 82.2 & 58.9 & 94.0 & 72.4 \\
            \midrule
            \midrule
            CAiDA + KD~\cite{hinton2015distilling} & 0.029 & 53.6 & 93.6 & 68.2 & 48.9 & 98.9 & 65.4 & 49.8 & 98.2 & 66.1 & 50.1 & 83.1 & 62.5 & 65.9 & 97.6 & 78.7 & 53.7 & 94.3 & 68.4 \\
            CAiDA + \textbf{SepRep-Net} & 0.032 & 57.1 & 94.0 & 71.0 & 56.4 & 99.2 & 71.9 & 55.9 & 98.8 & 71.4 & 56.8 & 85.7 & 68.3 & 74.2 & 98.3 & 84.6 & \textbf{60.1} & \textbf{95.2} & \textbf{73.7} \\
            \midrule
            CAiDA~\cite{dong2021confident} & 0.116 & 55.6 & 93.7 & 69.8 & 52.3 & 99.1 & 68.5 & 53.1 & 98.6 & 69.0 & 52.8 & 83.3 & 64.6 & 68.0 & 98.1 & 80.3 & 56.4 & 94.6 & 70.6 \\
            \midrule
            \midrule
            SHOT-ens + KD~\cite{hinton2015distilling} & 0.029 & 57.1 & 90.2  & 69.9 & 53.0 & 98.8 & 69.0 & 50.9 & 97.8 & 67.0 & 53.2 & 58.1 & 55.5 & 67.9 & 83.9 & 75.1& 56.4 & 85.8 & 68.1\\
            SHOT + \textbf{SepRep-Net}  & 0.032 & 67.6 & 95.8 & 79.3 & 63.1 & 98.6 & 77.0 & 70.7 & 97.6 & 82.0 &  71.2 & 82.9 & 76.6 &  75.9 & 93.1 & 83.6 & \textbf{69.7} & \textbf{93.6} & \textbf{79.9} \\
            \midrule
            SHOT-ens~\cite{liang2020shot}   &  0.116 &  60.3 &  90.4 &  72.3 &  56.7 & 98.9  &  72.1 &  60.8 & 97.7 &  75.0 &  54.9  & 58.3 &  56.5 &  68.9 & 83.9 &  75.6 &  60.3  & 85.8 &  70.8 \\
            \bottomrule
        \end{tabular}
    }
\end{table*}

\begin{table*}[!htbp]
    \centering
    \caption{Source Accuracy (S) (\%), Target Accuracy (T) (\%), H-Score (H), and Model Efficiency on DomainNet dataset. ResNet-101 is adopted in experiments. C, I, P, Q, R, S indicate different domains. (C: Clipart, I: Infograph, P: Painting, Q: Quickdraw, R: Real World, S: Sketch). The abbreviations in METHOD are the same as Table~\ref{tab: Acc-digit}.}
    \vspace{10pt}
    \resizebox{0.95\textwidth}{!}{
      \begin{tabular}{l|c|ccc|ccc|ccc|ccc|ccc|ccc|ccc}
        \toprule
        \multirow{2}{*}{METHOD}  & \multirow{2}{*}{FLOPS} & \multicolumn{3}{c|}{I,P,Q,R,S $\rightarrow$ C}      & \multicolumn{3}{c|}{C,P,Q,R,S $\rightarrow$ I}      & \multicolumn{3}{c|}{C,I,Q,R,S $\rightarrow$ P}      & \multicolumn{3}{c|}{C,I,P,R,S $\rightarrow$ Q}      & \multicolumn{3}{c|}{C,I,P,Q,S $\rightarrow$ R}       & \multicolumn{3}{c|}{C,I,P,Q,R $\rightarrow$ S}      & \multicolumn{3}{c}{Avg} \\
        & & S      & T      & H & S      & T      & H & S      & T      & H & S      & T      & H  & S      & T      & H & S      & T      & H & S      & T      & H\\
        \midrule
        DAN~\cite{Long15DAN}   & 39.2 & - & 39.1 & - & - & 11.4 & - & - & 33.3  & - & - & 16.2 & - & - & 42.1 & - & -& 29.7 & - & - & 28.6 & - \\
        DCTN~\cite{xu2018deep}    & 39.2 & -  & 48.6 & - & - & 23.4 & - & - & 48.8  & - & -  & 7.2 & - & - & 53.5 & - & - & 47.3 & - & - & 38.1 & - \\
        MCD~\cite{MCD2018}   & 39.2  & - & 54.3 & - & - & 22.2 & - & - & 45.7 & - & - & 7.6 & - & - & 58.4 & - & - & 43.5 & - & - & 38.6 & - \\
        M3SDA~\cite{peng2019moment}  & 39.2 & -  & 58.6 & - & - & 26.0 & - & - & 52.3  & - & - & 6.3 & - & -  & 62.7 & - & - & 49.5 & - & - & 42.5 & - \\
        \midrule
        \midrule
        DECISION + KD~\cite{hinton2015distilling} & 7.8 & 25.3  & 61.1  & 35.8  & 31.5  & 21.0 & 25.2 & 32.0  & 53.4  & 40.0 & 2.4  & 18.2 & 4.2 & 29.8  & 67.1 & 41.3 & 29.0 & 50.4 & 36.8 & 25.0  & 45.2 & 30.6   \\          
        DECISION + \textbf{SepRep-Net} & 7.8 & 33.1 & 62.0 & 43.2 & 34.9 & 22.3 & 27.2 & 37.4 & 55.0 & 44.5 & 3.3 & 18.5 & 5.6 & 33.4 & 68.1 & 44.8 & 33.6 & 51.8 & 40.8 & \textbf{29.3} & \textbf{46.3} & \textbf{35.9} \\ 
        \midrule
        DECISION~\cite{decision} & 39.2 & 31.4 & 61.5 & 41.6 & 34.3 & 21.6 & 26.5 & 36.2 & 54.6 & 43.5 & 3.1 & 18.9 & 5.3 & 32.4 & 67.5 & 43.8 & 32.8 & 51.0 & 39.9 & 28.4 & 45.9 & 35.1 \\
        \midrule
        \midrule
        CAiDA + KD~\cite{hinton2015distilling} & 7.8 & 25.1 & 61.3 & 35.6 & 29.8 & 21.8 & 25.2 & 32.2 & 54.6 & 40.5 & 2.5 & 19.1 & 4.4 & 27.6 & 67.8 & 39.2 & 30.0 & 50.9 & 37.8 & 24.5 & 45.9 & 32.0 \\
        CAiDA + \textbf{SepRep-Net} & 7.8 & 32.5 & 62.5 & 42.8 & 33.8 & 22.8 & 27.2 & 36.5 & 55.6 & 44.1 & 3.1 & 20.1 & 5.4 & 32.5 & 68.5 & 44.1 & 34.0 & 51.7 & 41.0 & \textbf{28.7} & \textbf{46.9} & \textbf{35.6} \\ 
        \midrule
        CAiDA~\cite{dong2021confident} & 39.2 & 29.8 & 61.9 & 40.2 & 32.8 & 22.2 & 26.5 & 34.6 & 55.0 & 42.5 & 3.0 & 19.3 & 5.2 & 29.3 & 68.1 & 41.0 & 31.3 & 51.2 & 38.8 & 26.8 & 46.3 & 33.9\\
        \midrule
        \midrule
        SHOT-ens + KD~\cite{hinton2015distilling} & 7.8 & 35.3 & 58.2 & 43.9 & 34.6 & 24.9 & 29.0 & 37.4 & 54.8 & 44.5 & 2.2 & 15.4 & 3.9 & 28.9 & 70.0 & 40.9 & 31.7 & 52.1 & 39.4 & 28.4 & 45.9 & 35.1\\
        SHOT + \textbf{SepRep-Net}  & 7.8 & 38.6 & 59.3 & 46.8 & 40.0 & 26.3 & 31.7 & 40.2 & 56.9 & 47.1 & 3.7 & 14.9 & 5.9 & 38.6 & 71.9 & 50.2 & 39.0 & 53.6 & 45.1 & \textbf{33.3} & \textbf{47.2}  & \textbf{39.0}  \\
        \midrule
        SHOT-ens~\cite{liang2020shot}  & 39.2  & 40.6 & 58.6 & 48.0& 37.2 & 25.2 & 30.0 & 41.1 & 55.3 & 47.2 & 3.0 & 15.3 & 5.0& 31.6 & 70.5 & 43.6 & 35.3 & 52.4 & 42.2 & 31.5 & 46.2 & 37.5\\
        \bottomrule
      \end{tabular}}
    \label{tab: Acc-DomainNet}
  \end{table*}

\begin{table*}[htbp]
    \centering
    \caption{Source Accuracy (S) (\%), Target Accuracy (T) (\%), H-Score (H), and Model Efficiency on Office-Home dataset. ResNet-50 is adopted in experiments. Ar, Cl, Pr, Rw indicate different domains (Ar: Art, Cl: Clipart, Pr: Product, Rw: Real World). Source-ens indicates the performance of model ensemble with all source models. The abbreviations in METHOD are the same as Table~\ref{tab: Acc-Office}.}
    \vspace{10pt}
    \label{tab: ana-src-ens}
    \small
    \resizebox*{\textwidth}{!}{
        \begin{tabular}{l|c|ccc|ccc|ccc|ccc|ccc}
            \toprule
            \multirow{2}{*}{METHOD}  & \multirow{2}{*}{FLOPS}   & \multicolumn{3}{c|}{Cl,Pr,Rw $\rightarrow$ Ar} & \multicolumn{3}{c|}{Ar,Pr,Rw $\rightarrow$ Cl} & \multicolumn{3}{c|}{Ar,Cl,Rw $\rightarrow$ Pr} & \multicolumn{3}{c|}{Ar,Cl,Pr $\rightarrow$ Rw} & \multicolumn{3}{c}{Avg}  \\                  
            &                        & S      & T      & H      & S      & T      & H      & S      & T      & H      & S      & T      & H   & S      & T      & H \\             
            \midrule
            Source-ens~\cite{he2016Resnet}  & 12.3 & 90.6 & 58.4 & 71.0 & 91.2 & 43.0 & 58.4 & 88.5 & 67.7 & 76.7 & 89.1 & 70.8 & 78.9 & \textbf{89.9} & 60.0 &72.0 \\
            SHOT-ens + KD~\cite{hinton2015distilling} & 4.1 & 69.0 & 72.1 & 70.5 & 65.6 & 59.0 & 62.1 & 67.9 & 82.7 & 74.6 & 66.5 & 82.9 & 73.8 & 67.3 & 74.2 & 70.6 \\
            SHOT + \textbf{SepRep-Net}  & 4.1 & 82.6 & 72.3 & 77.1 & 75.6 & 59.2 & 66.4 & 75.1 & 83.3 & 79.0 & 77.8 & 83.3 & 80.5 & 77.8 & \textbf{74.5} & \textbf{76.1}  \\
            \bottomrule
        \end{tabular}
    }
\end{table*}

\begin{table*}[ht]
    \centering
    \caption{Source Accuracy (S) (\%), Target Accuracy (T) (\%), H-Score (H), and Model Efficiency on Digit5. The backbone network in \cite{peng2019moment} is adopted to recognize digits. MM, MT, UP, SV, and SY indicate different domains (MM: MNIST-M, MT: MNIST, UP: USPS, SV: SVHN, SY: Synthetic Digits). $\rightarrow$ points to the target domain while the remaining domains serve as source domains. The abbreviations are the same as Table~\ref{tab: Acc-digit}.}
    \label{tab: Acc-digit-more}
    \footnotesize
    \vspace{10pt}
    \resizebox*{0.95\textwidth}{!}{
        \begin{tabular}{l|c|ccc|ccc|ccc|ccc|ccc|ccc}
            \toprule
            \multirow{2}{*}{METHOD} & \multirow{2}{*}{FLOPS} & \multicolumn{3}{c|}{ $\rightarrow$ MM} & \multicolumn{3}{c|}{$\rightarrow$ MT} & \multicolumn{3}{c|}{ $\rightarrow$ UP} & \multicolumn{3}{c|}{$\rightarrow$ SV} & \multicolumn{3}{c|}{$\rightarrow$ SY} & \multicolumn{3}{c}{Avg} \\
            &                        & S      & T      & H      & S      & T      & H      & S      & T      & H      & S      & T      & H   & S      & T      & H      & S      & T      & H\\
            \midrule
            BV-MSFDA + KD~\cite{hinton2015distilling} & 0.029 & 52.4 & 97.3 & 68.1 & 46.9 & 99.1 & 63.7 & 47.1 & 98.4 & 63.7 & 49.2 & 90.6 & 63.8 & 63.8 & 98.4 & 77.4 & 51.9 & 96.7 & 67.3 \\
            BV-MSFDA + \textbf{SepRep-Net} & 0.032 & 56.4 & 97.6 & 71.5 & 55.6 & 99.3 & 71.3 & 54.7 & 98.6 & 70.4 & 55.9 & 91.8 & 69.5 & 72.5 & 98.7 & 83.6 & \textbf{59.0} & \textbf{97.2} & \textbf{73.2}\\
            \midrule
            BV-MSFDA & 0.116 & 55.1 & 97.4 & 70.4 & 54.9 & 99.2 & 70.7 & 53.8 & 98.4 & 69.6 & 54.7 & 90.7 & 68.2 & 70.2 & 98.4 & 81.9 & 57.7 & 96.8 & 72.2\\
            \bottomrule
        \end{tabular}
    }
\end{table*}

\begin{table}[!htbp]
    \centering
    \caption{
        \textbf{Ablation Study} of different designs in SepRep-Net on Office-Home with ResNet-50 backbone. We take SHOT as the base method. H-score is used as the evaluation metric here. The meaning of $\rightarrow$ is the same as Table~\ref{tab: Acc-digit}. (Sep: Separation, Rep: Reparameterization, ReW: Importance Reweighting)
    }
    \vspace{10pt}
    \resizebox{0.48\textwidth}{!}{
        \begin{tabular}{lcccccc}
            \toprule
                                     &FLOPs   & $\rightarrow$Ar & 
                                        $\rightarrow$ Cl & 
                                        $\rightarrow$ Pr &
                                        $\rightarrow$ Rw & Avg           \\
            \midrule
            + Sep  &  12.3 & 73.4 & 65.0 & 77.5 & 77.9 & 73.5 \\
            + Sep (+ KD) & 4.1 & 70.2 & 62.3 & 74.1 & 76.5 & 70.8  \\ 
            + Sep + Rep & 4.1 & 73.4 & 65.0 & 77.5 & 77.9 & 73.5 \\
            \textbf{+ Sep + Rep + ReW (ours)} & 4.1 & \textbf{77.1} & \textbf{66.4} & \textbf{79.0} & \textbf{80.5} & \textbf{76.1}\\
            \bottomrule
        \end{tabular}
        }
        \vspace{-10pt}
        \label{tab:ablation}
\end{table}

We build SepRep-Net based on various baseline frameworks and evaluate the effectiveness, efficiency, and generalizability of SepRep-Net on different benchmarks.
Specifically, we first plug SepRep-Net into some typical methods, DECISION~\cite{decision}, SHOT~\cite{liang2020shot} and CAiDA~\cite{dong2021confident}, respectively. DECISION and CAiDA are designed for MSFDA tasks. SHOT is originally designed for SFDA, and it can be readily applied to MSFDA via model ensemble (SHOT-ens). We compare it with the vanilla results with multiple models and results with knowledge distillation~\cite{hinton2015distilling} (KD), a mainstream solution to enhance model efficiency. Following \cite{yang2021generalized}, we report the average accuracy on source domains $Acc_s$ ($\%$), the accuracy on target domain $Acc_t$ ($\%$) and the H-score $\frac{2 \cdot Acc_s \cdot Acc_t}{Acc_s + Acc_t}$ ($\%$) which jointly evaluates effectiveness and generalizability. On efficiency, we report the FLOPs of each method. The experiment results are reported in Table~\ref{tab: Acc-Office}, Table~\ref{tab: Acc-Office-home}, Table~\ref{tab: Acc-Office-caltech}, Table~\ref{tab: Acc-digit} and Table~\ref{tab: Acc-DomainNet}. 

In these tables, for MSFDA methods (DECISION and CAiDA), we report their performance according to their original paper.
For SFDA methods such as SHOT, to tackle the multi-source free scenario, we first adapt multiple models to the target domain via SHOT one by one.
SHOT-ens indicates the performance of model ensemble with all source models that are adapted via SHOT.

\paragraph{Office31} Table~\ref{tab: Acc-Office} shows that knowledge distillation (KD) is effective in enhancing efficiency and maintaining target performance, but it harms generalizability (obviously low source accuracy). On the contrary, reducing computational costs by half, SepRep-Net preserves significantly more generalizability than KD, as well as achieves better results in both source and target domains.

\paragraph{Office-Home and Office-Caltech} Table~\ref{tab: Acc-Office-home} and Table~\ref{tab: Acc-Office-caltech} demonstrate that when tackling three source models, our framework shows consistently higher performance than other methods in both source and target domains. We note that when compared with the vanilla methods, our method is still superior to them both in effectiveness and generalizability, with only $33\%$ computational costs of them. 

\paragraph{Digit5} Moreover, we further evaluate our method on Digit5, a harder dataset with more domains and larger domain gaps. The results in Table~\ref{tab: Acc-digit} prove that our method only requires less than $25\%$ of the computational costs, while consistently improving the original method with $100\%$ costs in both source and target accuracy. Here, we also consider previous works that have access to source data during being adapted to the target domain.
It is unfair to evaluate the source domain accuracy of these methods. Therefore, we only consider the accuracy of these methods in the target domain. Our method achieves a stronger target domain accuracy over these methods.

\paragraph{DomainNet} Finally, we take DomainNet, a larger dataset in domain adaptation, with obvious domain gaps across the six domains. Table~\ref{tab: Acc-DomainNet} shows that when compared with knowledge distillation, our method shows obvious improvements in effectiveness and generalizability under equal computational costs. For the vanilla methods, our method still surpasses them consistently, with only $20\%$ of the original computation costs.

\subsection{Analyses}




\begin{figure}[ht]
  \centering
  \subfigure[]{\label{fig:output-weight-1}\includegraphics[width=0.23\textwidth]{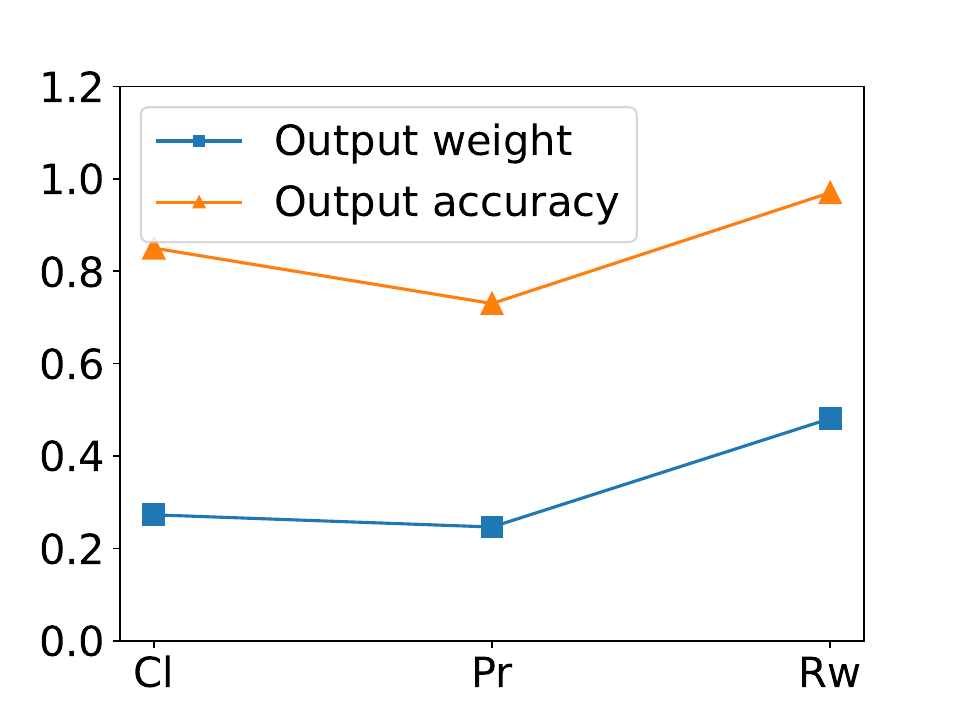}}
  \hfill
  \subfigure[]{\label{fig:output-weight-2}\includegraphics[width=0.23\textwidth]{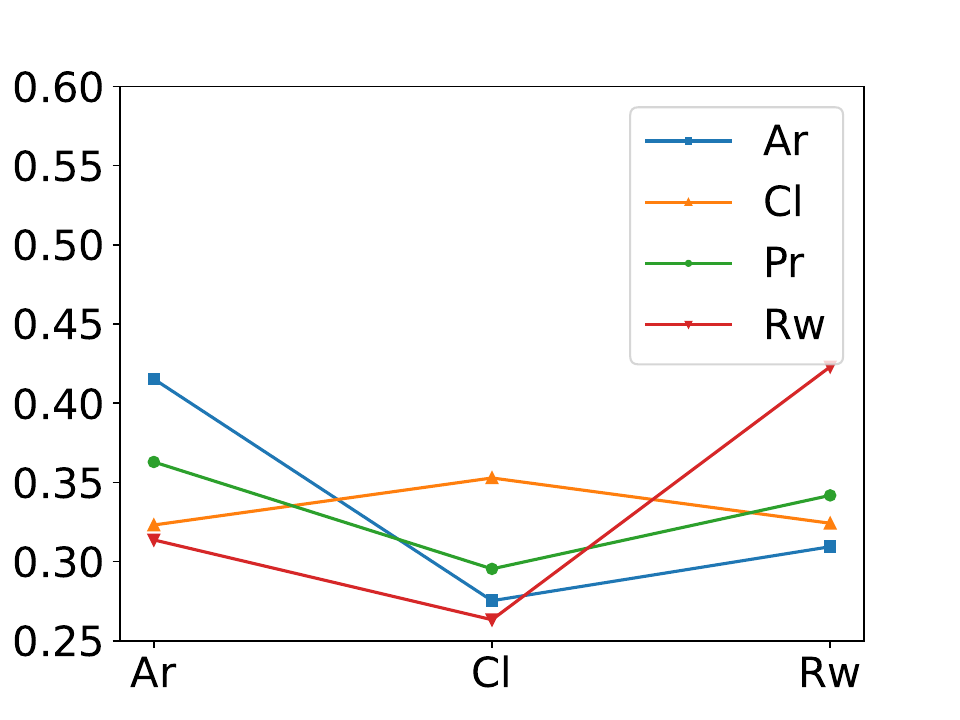}}
  \vspace{-10pt}
  \caption{\textbf{Output Weights} for different domains in Office-Home with ResNet-50 backbone when applying SepRep-Net to SHOT. (a) The output weights calculated in our framework are positively correlated to the classifier accuracy. (b) Our importance weights are adaptive to the input. Specifically, the importance weight of the input domain adaptively becomes larger than others, enhancing generalizability.}
  \label{fig:output-weight}
  \vspace{-10pt}
\end{figure}

\paragraph{More Analysis on Source Accuracy} The model accuracy on source domains is often taken as the metric to evaluate the model generalizability. For source accuracy, the ensemble results of multiple source models no doubt enjoys a strong performance. Therefore, it is interesting to compare our method with it. From Table~\ref{tab: ana-src-ens}, we can observe that ensembling multiple source models (Source-ens) shows strong performance in source domains, but poor accuracy in the target domain. The source accuracy gap between our method and Source-ens is much smaller than the traditional knowledge distillation, with similar computational costs. On the other hand, H-score validates that our method keeps the best trade-off between effectiveness and generalizability among these methods.

\paragraph{Results on More Method} We evaluate our method on a more recent method, BV-MSFDA~\cite{shen2023balancing}. As shown in Table~\ref{tab: Acc-digit-more}, our method cooperates with this recent method well.

\paragraph{Ablation Study} We investigate the components in SepRep-Net: Separation (Sep), Reparameterization (Rep), and Importance Reweighting (ReW). We mention that when taking separate pathways alone, we adopt traditional knowledge distillation on the model with multiple pathways, which makes the model size equal to our method for a fair comparison. H-score results are reported in Table~\ref{tab:ablation}, justifying that the effect of each part is indispensable.

\paragraph{Importance Reweighting}
We analyze the importance weights for ensembling outputs when the model is evaluated on different domains respectively. As shown in Figure~\ref{fig:output-weight}, our importance weight has an obviously positive correlation with the accuracy of each classifier on the unlabeled target domain. Moreover, our reweighting strategy is actually adapted to the input data during inference, leading to improved generalizability.


\section{Conclusion}
\label{sec:conclusion}
In this paper, we target on reassembling multiple existing models into a single model and adapting it to a novel target domain without accessing source data. Towards this problem, we propose SepRep-Net, a framework that forms multiple separate pathways during training and further merges them via reparameterization to facilitate inference. As a general approach, SepRep-Net is easy to be plugged into various methods. Extensive experiments prove that SepRep-Net consistently improves existing methods in effectiveness, efficiency, and generalizability. 

\paragraph{Impact Statements}
This paper proposes Multi-source Free Domain Adaptation with effectiveness, efficiency, and generalizability, a practical problem that is worth researching. For instance, to tackle medical problems, we may have multiple models trained on patient data collected from different hospitals. The data is no doubt inaccessible due to privacy concerns. With techniques in this paper, we can adapt them to new data, as well as maintain the performance on previous data. 



\bibliography{seprep}
\bibliographystyle{icml2024}

\newpage
\appendix
\onecolumn



\section{Appendix}
\paragraph{Dataset Details}
We perform extensive evaluation of SepRep-Net on five benchmark datasets, \ie, \textbf{Office-31}~\cite{Saenko10Office}, \textbf{Office-Home}~\cite{Venkateswara17Officehome}, \textbf{Digit5}~\cite{peng2019moment}, \textbf{Office-Caltech}~\cite{GFK}, and \textbf{DomainNet}~\cite{peng2019moment}.
\textbf{1) Office-31}~\cite{Saenko10Office} contains $4,652$ images from $31$ categories collected in the office environment, which forms three domains, \ie, Amazon (\textbf{A}), DSLR (\textbf{D}) and Webcam (\textbf{W}); \textbf{2) Office-Home}~\cite{Venkateswara17Officehome} consists of $15,500$ images, $65$ categories from $4$ domains, \ie, Art (\textbf{Ar}), Clipart (\textbf{Cl}), Product (\textbf{Pr}) and Real World (\textbf{Rw}). The domain gap in Office-Home is much larger than Office-31; \textbf{3) Digit5}~\cite{peng2019moment} is a benchmark dataset for cross-domain digit recognition, with $5$ domains, \ie, MNIST (\textbf{MT}), USPS (\textbf{UP}), SVHN (\textbf{SV}), MNIST-M (\textbf{MM}) and Synthetic Digits (\textbf{SY});
\textbf{4) Office-Caltech}~\cite{GFK} takes the common classes between Caltech-256 and Office-31. It has images from $10$ categories and $4$ domains, \ie, Amazon (\textbf{A}), Caltech (\textbf{C}), DSLR (\textbf{D}), and Webcam (\textbf{W});
\textbf{5) DomainNet}~\cite{peng2019moment} has 345 categories and 6 domains, \ie, Infograph (\textbf{I}), Clipart (\textbf{C}), Painting (\textbf{P}), Quickdraw (\textbf{Q}), Real (\textbf{R}) and Sketch (\textbf{S}).

\paragraph{Implementation Details} We take one domain as the target domain and view the remaining domains as source domains, which forms multiple tasks in each dataset. For real-world objects, we follow the conventions of image size and augmentation in \cite{decision,liang2020shot}. For digit datasets, we rescale images to $32 \times 32$ and convert gray-scale images to RGB. We obtain source models following the training scheme in \cite{decision,liang2020shot} to ensure a fair comparison.
The target model is trained with the SGD optimizer. The training schedule follows the settings in \cite{liang2020shot}. Moreover, the learning rate of the bottleneck and classifier is $10 \times$ larger than the feature extractor. The whole framework is trained end-to-end, with 1 GPU. Distributed optimization can adapted to accelerate training~\cite{wang2019distributed,zhou2019finite,wang2017distributed,zhou2018distributed}. 
For knowledge distillation, we adopt the vanilla knowledge distillation method proposed in \cite{hinton2015distilling}, since among existing methods~\cite{fitnets,takd,revkd,jin2023multi}, it is the most widely applied method in domain adaptation application. 
We implement our experiments with Pytorch, run each experiment $5$ times and report the average results. 

\begin{table}[htbp]
    \caption{
        \textbf{Comparison between different uncertainty metrics} for importance reweighting on Office-Home with ResNet-50 backbone. We take SHOT as the base method. H-score is used as the evaluation metric here. The meaning of $\rightarrow$ is the same as Table~\ref{tab: Acc-digit}. 
    }
    \vspace{10pt}
    \centering
    \resizebox{0.4\textwidth}{!}{
    
        \begin{tabular}{lccccc}
            \toprule
                        & $\rightarrow$ Ar & $\rightarrow$ Cl & $\rightarrow$ Pr & $\rightarrow$ Rw & Avg           \\
            \midrule
            Confidence     & 75.4 & 66.0 & 78.3 & 78.9 & 74.7        \\
            Margin         & 76.5 & 65.8 & 78.5 & 79.9 & 75.2         \\
            \textbf{Entropy (ours)} & \textbf{77.1} & \textbf{66.4} & \textbf{79.0} & \textbf{80.5} & \textbf{76.1} \\
            \bottomrule
        \end{tabular}
        \label{tab:uncertainty-metric}
    }
\end{table}

\paragraph{Comparison of Different Uncertainty Criteria}
In our method, we adopt entropy as the uncertainty-based criterion in our re-weighting strategy. There also exist various uncertainty metrics, such as confidence or margin. Here we give the comparison results of utilizing these different metrics to re-weight the multiple classifiers. Table~\ref{tab:uncertainty-metric} presents that our method performs stably with different uncertainty metrics, while the entropy metric enjoys the highest performance.

\end{document}